\documentclass[letterpaper, 10 pt, conference]{ieeeconf}  

\IEEEoverridecommandlockouts                              

\usepackage{amsfonts}
\usepackage{cite}
\usepackage{graphicx}
\usepackage{array}

\title{\LARGE \bf
An Energy-Saving Snake Locomotion Gait Policy Obtained Using Deep Reinforcement Learning
}

\author{\authorblockN{Yilang Liu}
\authorblockA{Department of Mechanical Engineering\\
Carnegie Mellon University\\
Pittsburgh, Pennsylvania 15213\\
Email: yilangl@andrew.cmu.edu}
\and
\authorblockN{Amir Barati Farimani}
\authorblockA{Department of Mechanical Engineering\\
Carnegie Mellon University\\
Pittsburgh, Pennsylvania 15213\\
Email: barati@cmu.edu}}

\begin{document}

\maketitle
\thispagestyle{empty}
\pagestyle{empty}

\begin{abstract}

Snake robots, comprised of sequentially connected joint actuators, have recently gained increasing attention in the industrial field, like life detection in narrow space. Such robots can navigate through the complex environment via the cooperation of multiple motors located on the backbone. 
However, controlling the robots in an unknown environment is challenging, and conventional control strategies can be energy inefficient or even fail to navigate to the destination. 
In this work, a snake locomotion gait policy is developed via deep reinforcement learning (DRL) for energy-efficient control. We apply proximal policy optimization (PPO) to each joint motor parameterized by angular velocity and the DRL agent learns the standard serpenoid curve at each timestep. The robot simulator and task environment are built upon \textit{PyBullet}. 
Comparing to conventional control strategies, the snake robots controlled by the trained PPO agent can achieve faster movement and more energy-efficient locomotion gait. This work demonstrates that DRL provides an energy-efficient solution for robot control. 

\end{abstract}

\section{Introduction}

Snake robot, inspired by the natural movement of snakes, has gained increasing attention in field robot domain. 
Different from traditional wheeled robots, snake robots, motored by multiple joint actuators, have more degrees of freedom. This enables them to move freely in the narrow environment, which provides great potential in life searching and rescuing\cite{robotImp}\cite{wang2020directional}\cite{6595317}\cite{1013176}. The earliest research on modeling and actuating snake robots start in 1946 when Gary explicitly describes the mechanism of the snake robots\cite{robotIntro}. Recent studies have formulated three major types of gaits for snake robots: lateral undulation, concertina locomotion, and sidewinding. Concertina locomotion\cite{concertina} is a cylindrical gait enabling snake robots to perform spatial motion around the cylinder. Sidewinding is a complex model that combines the horizontal and vertical body wave\cite{Astley6200}, thus giving the robots the ability to climb. The most common one is to describe continuous lateral undulation as serpenoid curve developed by Hirose\cite{serpenoidCurve}. Each joint is exerted sinusoidal bending and it propagates along the joints with a certain phase offset. Such mathematical equation sufficiently depicts the natural forward motion of snakes. However, all the models require complicated and tedious hand-tuned parameters to actuate the robots. The empirical tuning process can be more challenging and energy-consuming when the robot is working in a complicated or even unknown environment. In recent years, the development of reinforcement learning algorithms combined with deep neural networks introduces a framework that can achieve robust, accurate, and efficient control\cite{https://doi.org/10.1002/aisy.201900171}. The trained model can even surpass the performance of a human. By motoring the actions of a real human playing Atari games, researchers find that it outperforms all previous approaches on six of the games and surpasses a human expert on three of them\cite{mnih2013playing}. 

\begin{figure}[ht]
    \centering
    \includegraphics[width=3.4 in]{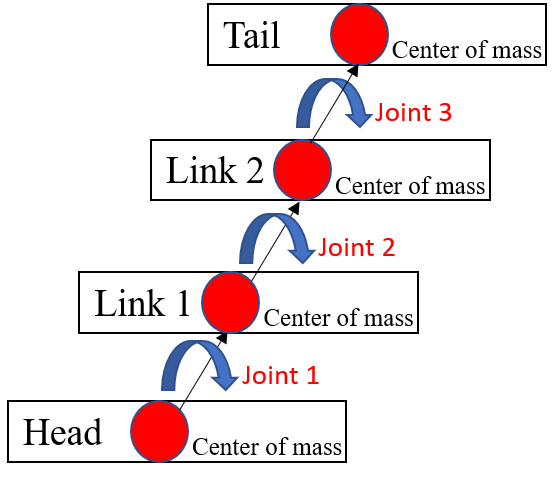}
    \caption{The structure of the snake robot}
    \label{images:snake}
\end{figure}

As for robot control, instead of using hand-tuning parameters, RL agents are capable of determining the best action options by themselves. Our contributions are summarized as follows. First, based on implementing DRL algorithms, we use simulation environments specifically \textit{PyBullet}\cite{coumans2019} to design a snake robot comprising step motors to test the performance of DRL designed controlling policy. By adding blocks to the environment and specifying the joints and links, we can create a customized robot same to the imported URDF model. The RL algorithms are carried out by Stable Baselines which are compatible with the current physics engine and are capable of implementing multiple RL algorithms based on OpenAI baselines\cite{stable-baselines}. The red circle represents the center of mass for each module and they are linked mechanically through the joints. The position and velocity of each joint can be obtained for estimating the current state of the robot. We then add torque on the joints sequentially to mimic the crawling behavior of the real snake as shown in Fig.\ref{images:snake}. The number of the joints are chosen to meet task requirement. Second, we implemented two RL algorithms to obtain observations and rewards from this simulation environment and apply actions to the robot based on the model predictions. Third, we designed a reward function, structure of the policy network, and the stopping criterion to achieve energy-efficient motion. They are tuned to lower the training time and precise action predictions. 

Overall, our results indicate 7.5\% faster speed and 38\% less energy consumed for each joint compared to the equation controller. Those results show that RL agents can carry out robot controlling by iterative training. The coupling of the simulation environment and RL algorithms enables users to easily develop control policies without design complex equations for every joint. Our simulated robot can also be a useful testing ground for various control strategies.

\section{Related Work}
Snake robots can perform complicated motion gaits by controlling multiple motors simultaneously. Researchers have been working on the implementation of deep reinforcement learning on adaptive and energy-efficient control\cite{Yu_2018}\cite{10.1177/0278364913495721} \cite{mnih2015human}. To safely operate the robot in diverse terrain conditions, they have to manually design and optimize the parameters of the functions to control the motors. At first, researchers designed the parameterized and scripted locomotion gaits to control the robot in a relatively simple function\cite{Tesch-2009-10250}. However, designing such a locomotion gait requires expensive objective function evaluations and time-consuming subsequent experiments. To optimize open-loop gaits parameters for snake robots, another implementation based on the response surface methodology is proposed\cite{6095076} but it still incapable of reducing energy efficiency. 

A few more studies focus on energy saving by applying machine learning to automate the parameter search. An evolutionary algorithm was adapted to learn high-quality walks. The results achieved 20\% improvement over best hand-tuned walks\cite{1389794}. To lower the computational cost, one group of competition teams used Powell's minimization method in automatic direction search and achieved 6\% faster than the previous had optimized gaits\cite{Hengst2001OmnidirectionalLF}\cite{Olave2003TheUR}\cite{article}.

Those controlling policies integrating machine learning algorithms in controlling the robot give a better performance than hand-tuned function. However, those algorithms do not take advantage of the previous learning experience and it usually converges to local optima. To better investigate the impact of prior knowledge on the current decision, Lizotte presented a Bayesian approach based on Gaussian process regression which addressed the expensive gait evaluations\cite{10.5555/1625275.1625428}. The analysis of Bayesian optimization in different configurations was also conducted and showed promising results\cite{6907117}. Recent researches tried to implement reinforcement learning in robot control without knowing the accurate model and prior knowledge of the environment\cite{Cully_2015}. More importantly, researchers find that RL based algorithms can facilitate the control of bio-inspired robots with consisting of high DOF such as Hexapod Robot\cite{10.3389/fnbot.2021.627157}\cite{phdthesis}\cite{lele2020learning},quad-rotor drone\cite{RAMEZANIDOORAKI2021103671}\cite{koch2018reinforcement}\cite{8453468}, bipedal robot\cite{6990997}\cite{8793627}\cite{castillo2019hybrid}. As more advanced algorithms are developed, the agents can control the robot to handle complicated tasks\cite{rajeswaran2018learning}\cite{long2018optimally}. Using hierarchical deep reinforcement learning, Peng indicated that DRL is capable of navigating through static or dynamic obstacles\cite{10.1145/3072959.3073602}. Moreover, by conducting real-world experiments with DRL, Petar\cite{10.1007/s10514-018-9697-6} achieved a significant 18\% reduction in the electric energy consumption and some of the models can even be adapted to learn robust control policies capable of imitating a broad range of example motion clips\cite{Peng_2018}. In this work, our model takes the advantage of DRL and applies it to learn snake locomotion gaits in a simulated environment.

\section{Method}
\begin{figure}
    \centering
    \includegraphics[width=3.4 in]{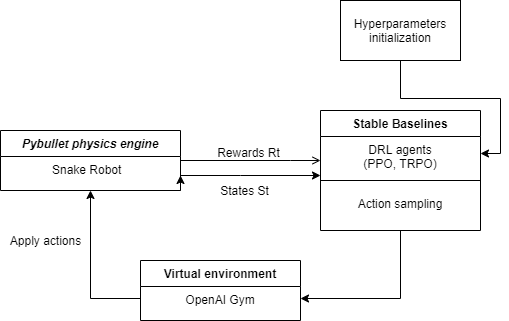}
    \caption{Flowchart of the RL training}
    \label{fig:flowchart}
\end{figure}

In this work, we applied proximal policy optimization (PPO) in the simulation to learn the gait. PPO is a policy gradient method that will optimize the "surrogate" objective function using stochastic gradient descent (SGD). Comparing to trust region policy optimization (TRPO), we choose PPO because it makes sure the policy does not go far from the old policy by clipping the probability ratio. First, we initialize the weights and parameters of RL agents. During the training process, the robot gives its current state and reward to RL agents. The agent will apply the next actions to the robot from its policy network and adjust weights based on the rewards. After the agent is trained, a complete simulation is performed and the trajectory is recorded for each timestep until the robot reaches the goal. The flowchart for the overall training is shown in Fig.\ref{fig:flowchart}. In the real world, the robot should have modulated joints that can give the robot capabilities suiting various tasks. To better discover which number of the joints are the most energy-efficient. The separated simulations are conducted based on the different joints of the robot. The energy consumption per joint will be recorded throughout the training to find the optimum joints suitable for the snake robot. 

\subsection{Proximal Policy Optimization (PPO)}
The main idea of PPO is to add a constraint to a surrogate objective function and using SGD to update the policy. PPO falls into the category of policy gradient algorithm, which uses the gradient method to directly update the policy rather than updating from the value function. The gradient estimator is given in Eq.~\ref{estimator}
\begin{equation}
    \hat{g}=\hat{\mathbb{E}}[{\nabla}_\theta{log\pi_\theta(a_t|s_t,\theta)\hat{A_t}}]
    \label{estimator}
\end{equation}
where $\pi_\theta$ is a stochastic policy at each timestep $t$, and $\hat{A_t}$ is the advantage estimates. The clipped surrogate objective in PPO is an alternative for the KL constraint in TRPO\cite{wu2019investigation}, which is defined in Eq.~\ref{Clip}:
\begin{equation}
L^{CLIP}(\theta)=\hat{\mathbb{E}}_t[min(r_t(\theta)\hat{A}_t,clip(r_t(\theta),1-\epsilon,1+\epsilon)\hat{A}_t)]
\label{Clip}
\end{equation}
where the probability ratio $r_t(\theta)$ is in Eq.~\ref{rt}: 
\begin{equation}
r_t(\theta)=\frac{\pi_\theta(a_t|s_t)}{\pi_{old}(a_t|s_t)}
\label{rt}
\end{equation}
$\hat{A}_t$ is the estimated advantage funtion at timestep $t$. And it is expressed in Eq.~\ref{Ahat}
\begin{equation}
\hat{A}_t=\sum_{k=0}^{\infty}\gamma^kR_{t+k+1}-V_t.  
\label{Ahat}
\end{equation}
If $\hat{A}_t$ is positive, it means the actions agent took is better than expected, so the policy gradient will be positive and increase the probability of the actions. Then we use $\epsilon$ to prevent the gradient update $r_t(\theta)$ from moving out of the interval $(1-\epsilon,1+\epsilon)$. 

\subsection{Snake Robot Simulator}
\begin{figure}[ht]
    \centering
    \includegraphics[width=3.4 in]{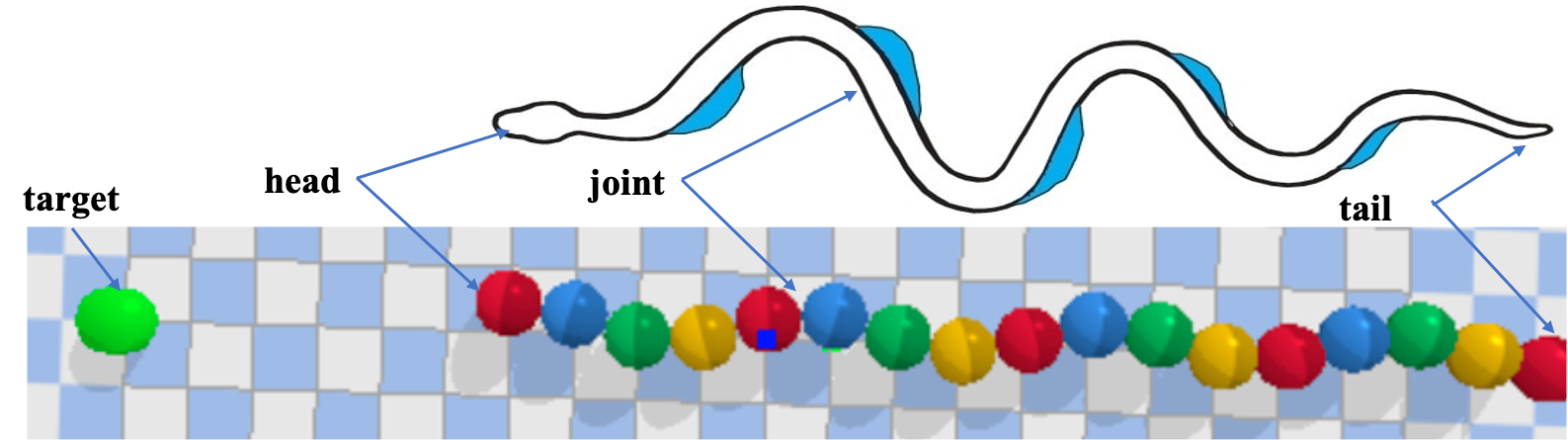}
    \caption{The snake robot simulator in \textit{PyBullet} simulation environment. The left green object is the cylindrical target placed in front of the robot. The coordinate of the target is [0,-10]. For the right side is the snake robot and it generates serpenoid curve for each joint to propel the robot moving forward.}
    \label{images:model}
\end{figure}
We construct the snake robot model with 17 joints and each joint links with two spheres that can rotate along the z-axis plane. To simulate the real ground, the anisotropic friction is set to [1, 0,01, 0.01].  After applying standard gravitational force to the robot, the snake robot can move toward the target by rotating the joints. To simplify the model, the model starts from the function in Eq.~\ref{Motion} to control the joint.
\begin{equation}
    \theta_i(t)=A\sin({\omega}t-(i-1)\phi)
    \label{Motion}
\end{equation}
The joint angle of the $i^{th}$ joint along the x-axis is the sin wave with $-(i-1)\phi$ offset. $A$ is the amplitude that controls the maximum moving range for each timestep $t$. $\omega$ is the movement speed of the joint and it determines the frequency of the movement. In this model, each timestep is $1/30$ second. In \textit{PyBullet} environment, we choose position control to motors with fixed force 10 $N$. Our learning objective is to let the agent take different values for $\omega$ and learn from the experience. Notice that there is no subscript in $\omega$ meaning all joints have the same moving speed. For the general structure of our reinforcement learning model. The action space in this work is the moving speed of the robot. As shown in Table \ref{tab:ob}, the observation space consists of three major parts including position, orientation, and velocity. After the agent receives the observations, it will pick an action for the next timestep.
\begin{table}
\renewcommand{\arraystretch}{1.3}
\caption{Observation Space in Snake Robot Environment}
\label{tab:ob}
\centering
\begin{tabular}{c||m{2.5 in}}
\hline
         Dimension & Observation Description \\
         \hline 
         0 & the Cartesian X coordinate of front head position on the surface \\
         \hline
         1 & the Cartesian Y coordinate of front head position on the surface \\
          \hline 
         2 & the sin value of front head orientation angle  \\
         \hline
         3 & the cosine value of front head orientation angle\\
          \hline 
         4 & the Cartesian X coordinate of centroid on the surface \\
         \hline
         5 & the Cartesian Y coordinate of centroid on the surface \\
          \hline 
         6 & the sin value of centroid orientation angle \\
         \hline
         7 & the cosine value of centroid orientation angle\\
          \hline 
         8 & the velocity value of centroid \\
         \hline
\end{tabular}
\end{table}

To make sure the policy updates toward minimum energy consumption while maintaining pure forward motion, the reward function is designed based on both the velocity and position of the snake robot. The reward function for the snake robot environment is expressed in Eq.\ref{Reward}.
\begin{equation}
    R_{reward} = max(X_{distance},0)+v_{velocity}-1
    \label{Reward}
\end{equation}
$X_{distance}$ is the distance traveled from timestep $t-1$ to $t$. $v_{velocity}$ is the forward velocity of the centroid. If the snake robot reaches the goal, the reward is set to 100. Any other state in which the robot is in will have a constant penalty of -1. In each episode, the robot and target will be reset to a fixed position, then the position to target and velocity are calculated. If the robot moves backward, the difference between two consecutive distances as well as velocity is a negative value. The first term in reward will be zero and the total reward is negative. 
The energy efficiency for individual joints can be calculated by integrating the trajectory over time and divided by the total time elapsed. An individual joint will consume the energy shown in Eq.\ref{individual energy}. For $k$ joints, the total energy efficiency $q$ per timestep $t$ is the multiplication of each joint's angular velocity $\dot{\phi_i}$ and torque $\tau$ from $0$ to $T$ divided by total time $T$. In this equation, noticing that each joint will have its unique motion pattern during the simulation meaning the angular velocity of each joint will be different.
\begin{equation}
    q_{k}= \frac{1}{T} \int_{0}^{T} \tau_k \dot{\phi_k}\,dt
    \label{individual energy}
\end{equation}
 However, this will not be able to tell the system behavior since every joint will have unique movement. One of the ways to evaluate the performance of the RL-controlled robot is to obtain the total energy consumed per timestep. To account for the total energy consumed during the simulation. In this work, the energy consumption is calculated by summing the individual joint's trajectory and divided by total timesteps after reaching the goal. The way to compare the energy efficiency for two gaits is shown in Eq.\ref{energy}
\begin{equation}
    q = \sum_{i=0}^{k} q_{k}
    \label{energy}
\end{equation}
For PPO architecture, the hyperparameters we used for the agent are hand-tuned to achieve faster results. They are shown in Table \ref{tab:PPO}.
\begin{table}
\renewcommand{\arraystretch}{1.3}
\caption{PPO hyperparameters for snake robot simulation}
\label{tab:PPO}
\centering
\begin{tabular}{c||c}
\hline
         Hyperparameter & Value \\
        \hline
         Total timesteps & 2e4 \\
         Discount factor($\gamma$) & 0.95\\
         Clip range ($\epsilon$) & 0.2\\
         GAE ($\lambda$) & 0.95\\
         VF coefficient & 0.5\\
         Number of epoch & 20\\
         Batch size & 5e4\\
         Learning rate (Adam) & 0.0002\\
    \end{tabular}
\end{table}

We also trained a TRPO agent for comparison, the structure of actor and critic is the same as PPO with minor changes in hyperparameters. The hyperparameters for TRPO are listed in Table \ref{tab:TRPO}.

\begin{table}
\renewcommand{\arraystretch}{1.3}
\caption{TRPO hyperparameters for snake robot simulation}
\label{tab:TRPO}
\centering
\begin{tabular}{c||c}
\hline
         Hyperparameter & Value \\
        \hline
         Total timesteps & 2e4 \\
         Timestep per batch & 2e3\\
         Discount factor($\gamma$) & 0.99\\
         GAE ($\lambda$) & 0.98\\
         KL loss threshold & 0.01\\
         Number of epoch & 20\\
    \end{tabular}
\end{table}

Since PPO and TRPO both use the Actor-Critic method. the agent needs two function approximators to estimate the value and policy function. Considering the action space is continuous, the network architectures are designed to have three hide layers. The Architecture of the actor and critic are given in Table \ref{tab:actor} and \ref{tab:critic}.
\begin{table}
\renewcommand{\arraystretch}{1.3}
\caption{The architecture of the actor}
\label{tab:actor}
\centering
\begin{tabular}{c||c||c}
\hline
        Layer & number of nodes & Activation \\
        \hline
         Fully connected & 100 & Tanh\\
         Fully connected & 50 & Tanh\\
         Fully connected & 25 & Tanh \\
         Output & 9 (dimension of the action space) & Tanh\\
    \end{tabular}
\end{table}

\begin{table}
\renewcommand{\arraystretch}{1.3}
\caption{The architecture of the critic}
\label{tab:critic}
\centering
\begin{tabular}{c||c||c}
        Layer & number of nodes & Activation \\
        \hline
         Fully connected & 100 & Tanh\\
         Fully connected & 50 & Tanh\\
         Fully connected & 25 & Tanh \\
         Output & 1  & Linear\\
    \end{tabular}
\end{table}

\section{Results}
\begin{figure}[ht]
    \centering
    \includegraphics[width=2.5in]{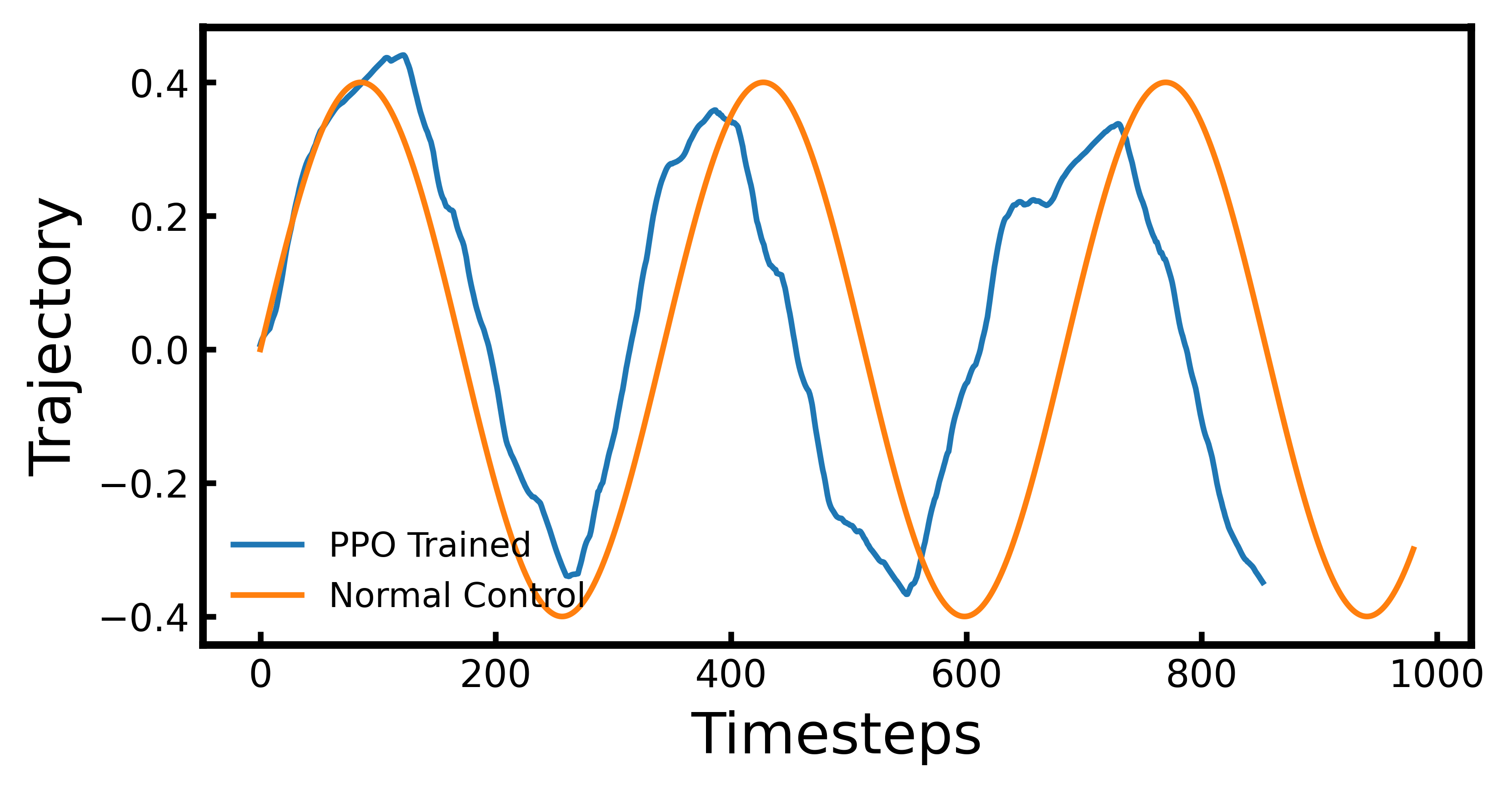}
    \caption{Trajectories of PPO gait and normal control gait}
    \label{fig:PPO_comparison}
\end{figure}

\textbf{Comparison of PPO and TRPO with a fixed number of joints.} In this work, we fix the target at 10 meters in front of the snake head and give a large penalty if the snake's centroid derives from forwarding motion. After setting up the environment, we find both the PPO and TRPO have successfully optimized the gait while maintaining forward motion. To evaluate the training results, the cumulative reward plots are generated. In Fig.\ref{fig:PPO_comparison}, We compare the learned gait with standard serpenoid control policy. Based on the model, other joints will follow the same pattern with a different phase shift. Then we can calculate the energy consumption and crawling velocity after reaching the target position. 

For learned gait in the blue line, the trajectories of the joints follow a similar sinusoidal wave forming a lateral undulation on the robot's backbone. Based on the two gaits comparison in Fig.\ref{fig:PPO_comparison}, the energy efficiency can be calculated using Eq.\ref{energy}. In the above graph, the snake robot controlled by the PPO agent takes 28.2 seconds, whereas normal control spends 33 seconds to arrive. During the simulation, each timestep is $1/30$ second. We found the energy consumed by learned gait is 0.152 $W$ and the hand-tuned control consumes 0.247 $W$. With the same force applied to each joint, the robot using learned gait has more energy efficient than normal control. Considering the time used for a different controller is different, the velocity of the learned snake robot is 0.35 $m/s$ which is faster than the standard policy of 0.3 $m/s$. Based on the observation of the robot, the PPO agent in this simulation is proved to perform more efficiently than the equation controller. 

\begin{figure}[ht]
    \centering
    \includegraphics[width=2.5in]{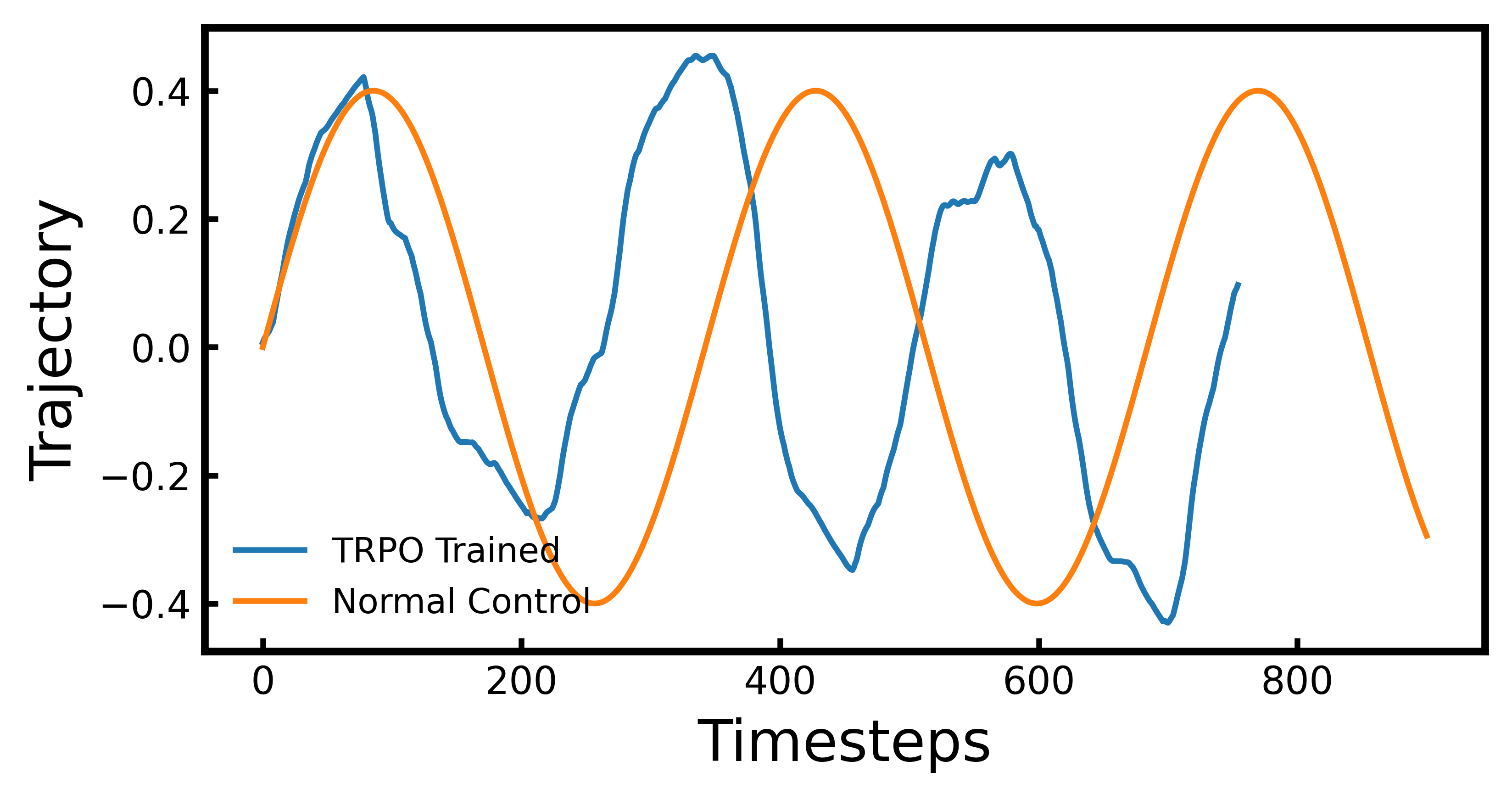}
    \caption{Trajectories of TRPO gait and normal control gait}
    \label{fig:TRPO_comparison}
\end{figure}

To make a comparison with PPO, we designed a TRPO agent to control the snake robot under the same environmental condition as shown in Fig.\ref{fig:TRPO_comparison}. 

Initially, the agent can follow the moving pattern. Compared to the normal control trajectory, the agent accelerates oscillation frequency as the simulation runs forward. The robot controlled by the TRPO agent takes 26 seconds. After integrating the area under the TRPO controlled trajectory and divided over time, the energy consumed is 0.201 $W$. Compared to PPO, even TRPO agent uses less time to reach the target position, it consumes more energy for each timestep.

To evaluate the training success, the accumulated reward after each episode during the training process is monitored and recorded.
\begin{figure}[ht]
    \centering
    \includegraphics[width=2.5in]{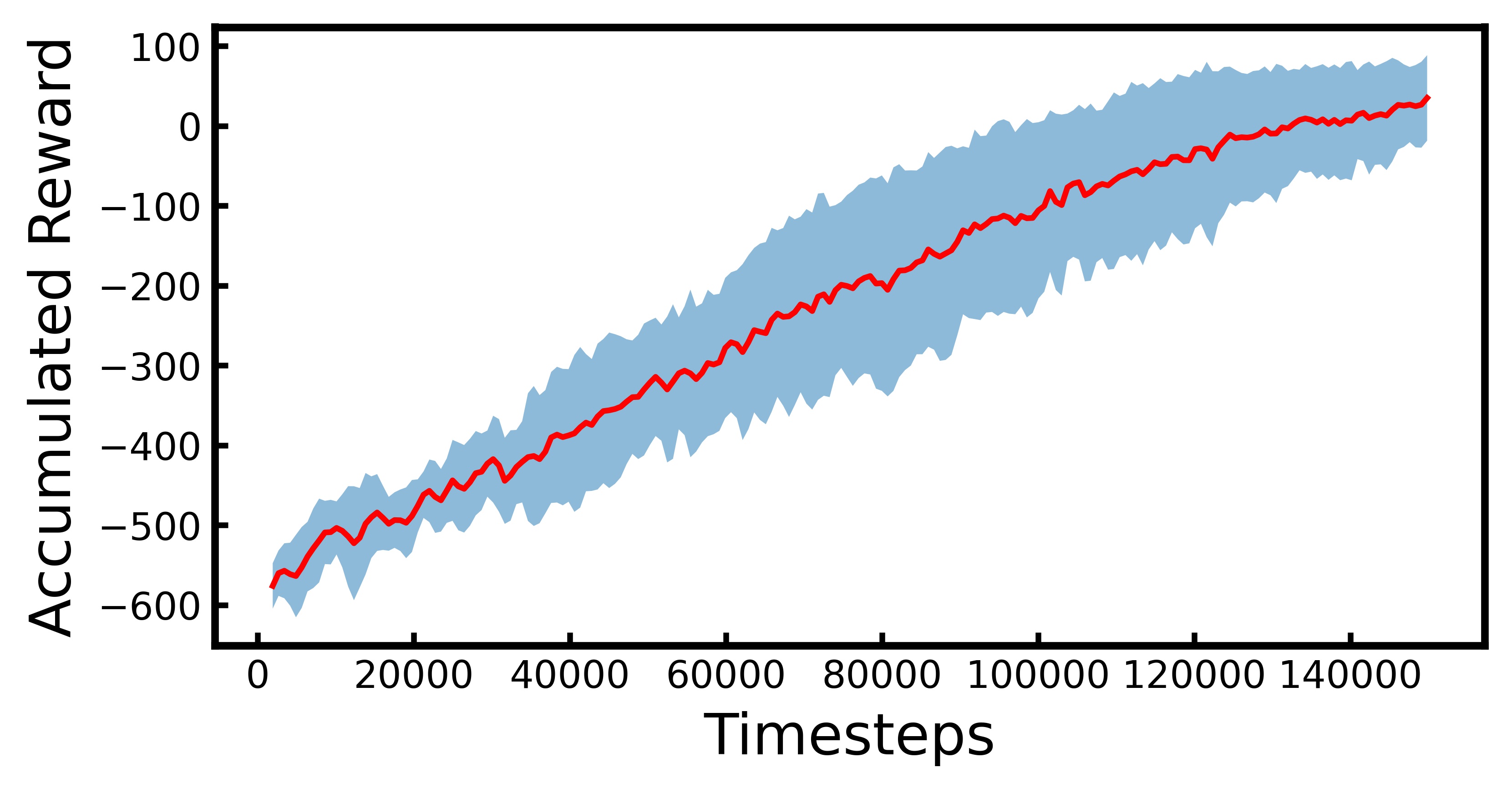}
    \caption{Accumulative reward versus timestep for PPO agent}
    \label{fig:rewardppo}
\end{figure}

\begin{figure}[ht]
    \centering
    \includegraphics[width=2.5in]{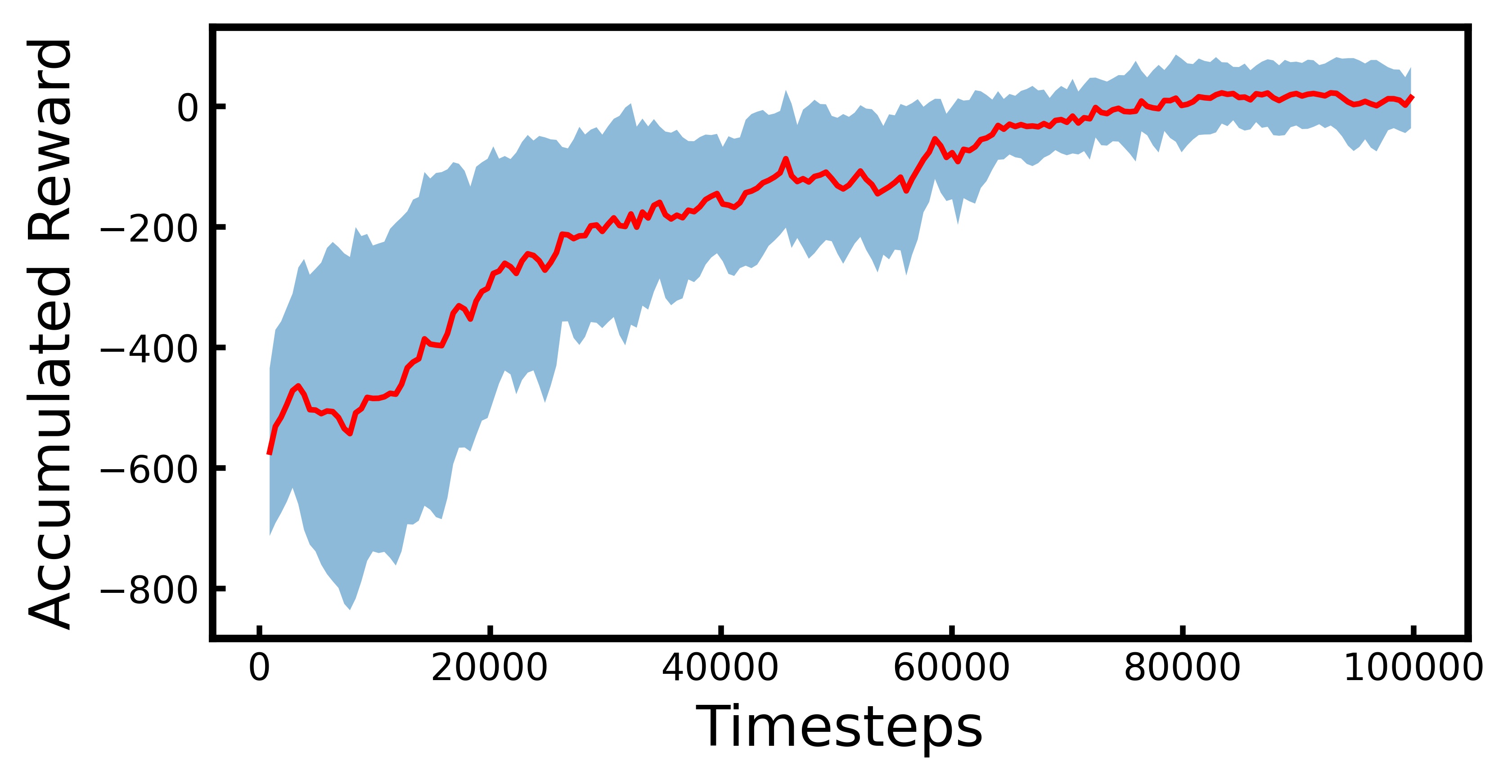}
    \caption{Accumulative reward versus timestep for TRPO agent}
    \label{fig:rewardtrpo}
\end{figure}

As shown in Fig.\ref{fig:rewardppo} and Fig.\ref{fig:rewardtrpo}, The total timesteps used for training PPO agent is 145000 and 100000 for TRPO agent. For TRPO and PPO agents, ten trials were conducted and their average reward versus timesteps are recorded in the red line. The shaded blue areas are the standard deviation during the training process. Noticing that more data points are recorded over time, the snake robot moves faster than in the previous episode. In the initial phase, the robot will take more time to wander around finding the way to reach the goal, which means fewer data points are recorded at first. As the training continues, the robot will spend less time approaching the goal and more data points will be recorded. Both DRL agents progressively approach 100 which is the reward value if the robot reaches the designed position. Compared to PPO, the TRPO agent takes less time but it can get results worse than the previous episode. PPO agent gives more consistent simulation results during the training. 

\begin{figure}
    \centering
    \includegraphics[width=2.5in]{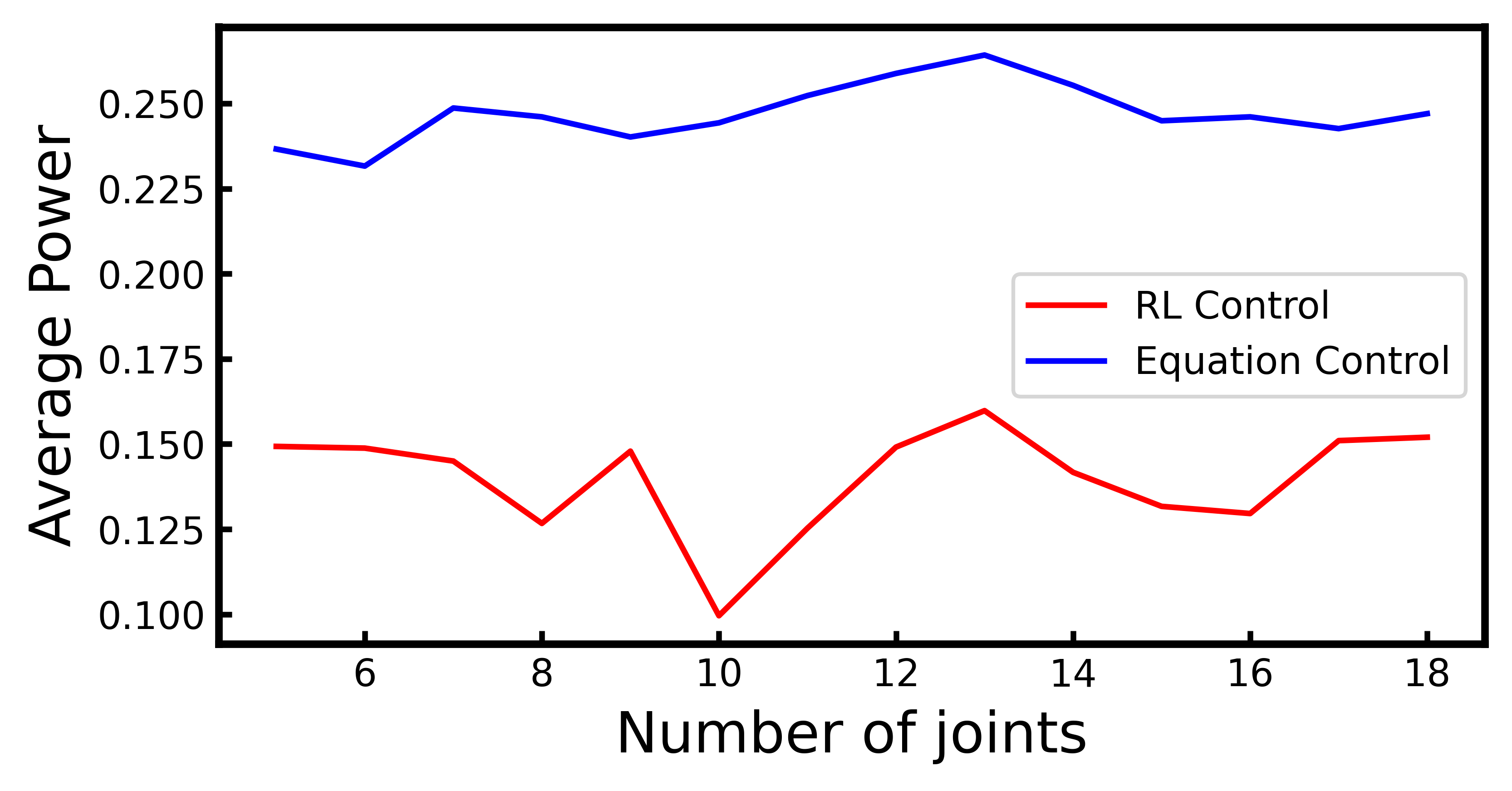}
    \caption{Power consumption with varied number of joints}
    \label{fig:myjoints}
\end{figure}

\textbf{Comparison of energy consumption per joint with the varied number of joints.} The number of the joint will affect the geometric shape of the robot as well as its physical properties. To determine the optimum number of locomotion joints, we use a PPO controller to let the robot run various motion modules while maintaining the same environment. The RL agents will try to optimize the control policy with the different number of locomotion joints. To evaluate the overall performance of the robot, the individual joint power will be recorded using Eq.\ref{individual energy}. Then, the average energy will be calculated by summing all individual power and take the average of it as shown in Eq.\ref{average energy}. 
\begin{equation}
    q_{average} = \frac{1}{K} \sum_{i=0}^{k} q_{k}
    \label{average energy}
\end{equation}
We start from 5 joints to 18 joints and conduct separated training for each trail. Fig.\ref{fig:myjoints} shows the results of how the joints will affect the general energy consumption of the robot. The blue line is the average power from equation controller which is the same throughout the trials. The red line is the average power from RL controller specifically PPO controller. The hand-tuned controller will generally have less energy efficiency according to the figure. Moreover, since the equation has the same control policy for different structures of the robot, the general energy consumption remains at the same level of performance. Comparing with the equation controller, we find that the PPO controller has a more flexible control over the robot and it has maximum energy efficiency when the robot has 10 joints. 

\section{Conclusion}
In this work, we develop an energy-efficient gait for snake robots based on deep reinforcement learning algorithms, specifically PPO and TRPO. Comparing both algorithms, we believe PPO gives more consistent and energy-efficient results. The learned gait is shown to achieve more sophisticated control than the existing equation controller while lowering energy consumption. The cumulative reward plot verifies the training converges after 100000 timesteps. 

The robot controlled by DRL trained agent has energy consumption lowered by 38\%, and crawling velocity increased by 7.5\%, comparing to conventional control strategies. 
The snake robot model built upon the gym environment, can serve as a benchmark for various DRL algorithms and allow people to customize the interaction with the environment. Currently, the robot only moves 10 meters in the simulation, the overall energy saving will be viable when putting it in longer trials. Our future work will include expanding the action space so that agents can obtain more control over the robot to achieve a better control policy.

\addtolength{\textheight}{-12cm}   

  
\bibliographystyle{IEEEtran}
\bibliography{IEEEabrv, reference}

\begin{thebibliography}{10}
\providecommand{\url}[1]{#1}
\csname url@samestyle\endcsname
\providecommand{\newblock}{\relax}
\providecommand{\bibinfo}[2]{#2}
\providecommand{\BIBentrySTDinterwordspacing}{\spaceskip=0pt\relax}
\providecommand{\BIBentryALTinterwordstretchfactor}{4}
\providecommand{\BIBentryALTinterwordspacing}{\spaceskip=\fontdimen2\font plus
\BIBentryALTinterwordstretchfactor\fontdimen3\font minus
  \fontdimen4\font\relax}
\providecommand{\BIBforeignlanguage}[2]{{%
\expandafter\ifx\csname l@#1\endcsname\relax
\typeout{** WARNING: IEEEtran.bst: No hyphenation pattern has been}%
\typeout{** loaded for the language `#1'. Using the pattern for}%
\typeout{** the default language instead.}%
\else
\language=\csname l@#1\endcsname
\fi
#2}}
\providecommand{\BIBdecl}{\relax}
\BIBdecl

\bibitem{robotImp}
M.~Tescha, K.~Lipkin, I.~Brown, R.~Hatton, A.~Peck, J.~Rembisz, and H.~Choset,
  ``Parameterized and scripted gaits for modular snake robots,'' \emph{Advanced
  Robotics}, pp. 1131--1158, 2009.

\bibitem{wang2020directional}
T.~Wang, J.~Whitman, M.~Travers, and H.~Choset, ``Directional compliance in
  obstacle-aided navigation for snake robots,'' 2020.

\bibitem{6595317}
M.~{Moattari} and M.~A. {Bagherzadeh}, ``Flexible snake robot: Design and
  implementation,'' in \emph{2013 3rd Joint Conference of AI Robotics and 5th
  RoboCup Iran Open International Symposium}, 2013, pp. 1--5.

\bibitem{1013176}
{Shugen Ma}, H.~{Araya}, and {Li Li}, ``Development of a creeping
  snake-robot,'' in \emph{Proceedings 2001 IEEE International Symposium on
  Computational Intelligence in Robotics and Automation (Cat. No.01EX515)},
  2001, pp. 77--82.

\bibitem{robotIntro}
J.~GRAY, ``The mechanism of locomotion in snakes,'' \emph{Jornal of
  Experimental Biology}, pp. 101--120, 1946.

\bibitem{concertina}
C.~Tang, X.~Shu, D.~Meng, and G.~Zhou, ``Arboreal concertina locomotion of
  snake robots on cylinders,'' \emph{International Journal of Advanced Robotic
  Systems}, vol.~14, p. 172988141774844, 11 2017.

\bibitem{Astley6200}
\BIBentryALTinterwordspacing
H.~C. Astley, C.~Gong, J.~Dai, M.~Travers, M.~M. Serrano, P.~A. Vela,
  H.~Choset, J.~R. Mendelson, D.~L. Hu, and D.~I. Goldman, ``Modulation of
  orthogonal body waves enables high maneuverability in sidewinding
  locomotion,'' \emph{Proceedings of the National Academy of Sciences}, vol.
  112, no.~19, pp. 6200--6205, 2015. [Online]. Available:
  \url{https://www.pnas.org/content/112/19/6200}
\BIBentrySTDinterwordspacing

\bibitem{serpenoidCurve}
S.~Hirose, \emph{Biologically Inspired Robots: Serpentile Locomotors and
  Manipulators}.\hskip 1em plus 0.5em minus 0.4em\relax Oxford University
  Press, 1993, vol. 240.

\bibitem{https://doi.org/10.1002/aisy.201900171}
\BIBentryALTinterwordspacing
K.~Chin, T.~Hellebrekers, and C.~Majidi, ``Machine learning for soft robotic
  sensing and control,'' \emph{Advanced Intelligent Systems}, vol.~2, no.~6, p.
  1900171, 2020. [Online]. Available:
  \url{https://onlinelibrary.wiley.com/doi/abs/10.1002/aisy.201900171}
\BIBentrySTDinterwordspacing

\bibitem{mnih2013playing}
V.~Mnih, K.~Kavukcuoglu, D.~Silver, A.~Graves, I.~Antonoglou, D.~Wierstra, and
  M.~Riedmiller, ``Playing atari with deep reinforcement learning,'' 2013.

\bibitem{coumans2019}
E.~Coumans and Y.~Bai, ``Pybullet, a python module for physics simulation for
  games, robotics and machine learning,'' \url{http://pybullet.org},
  2016--2019.

\bibitem{stable-baselines}
A.~Hill, A.~Raffin, M.~Ernestus, A.~Gleave, A.~Kanervisto, R.~Traore,
  P.~Dhariwal, C.~Hesse, O.~Klimov, A.~Nichol, M.~Plappert, A.~Radford,
  J.~Schulman, S.~Sidor, and Y.~Wu, ``Stable baselines,''
  \url{https://github.com/hill-a/stable-baselines}, 2018.

\bibitem{Yu_2018}
\BIBentryALTinterwordspacing
W.~Yu, G.~Turk, and C.~K. Liu, ``Learning symmetric and low-energy
  locomotion,'' \emph{ACM Transactions on Graphics}, vol.~37, no.~4, p. 1–12,
  Aug 2018. [Online]. Available:
  \url{http://dx.doi.org/10.1145/3197517.3201397}
\BIBentrySTDinterwordspacing

\bibitem{10.1177/0278364913495721}
\BIBentryALTinterwordspacing
J.~Kober, J.~A. Bagnell, and J.~Peters, ``Reinforcement learning in robotics: A
  survey,'' \emph{Int. J. Rob. Res.}, vol.~32, no.~11, p. 1238–1274, Sep.
  2013. [Online]. Available: \url{https://doi.org/10.1177/0278364913495721}
\BIBentrySTDinterwordspacing

\bibitem{mnih2015human}
V.~Mnih, K.~Kavukcuoglu, D.~Silver, A.~A. Rusu, J.~Veness, M.~G. Bellemare,
  A.~Graves, M.~Riedmiller, A.~K. Fidjeland, G.~Ostrovski \emph{et~al.},
  ``Human-level control through deep reinforcement learning,'' \emph{Nature},
  vol. 518, no. 7540, pp. 529--533, 2015.

\bibitem{Tesch-2009-10250}
M.~Tesch, K.~Lipkin, I.~Brown, R.~Hatton, A.~Peck, J.~M. Rembisz, and
  H.~Choset, ``Parameterized and scripted gaits for modular snake robots,''
  \emph{Advanced Robotics}, vol.~23, no.~9, pp. 1131 -- 1158, June 2009.

\bibitem{6095076}
M.~{Tesch}, J.~{Schneider}, and H.~{Choset}, ``Using response surfaces and
  expected improvement to optimize snake robot gait parameters,'' in \emph{2011
  IEEE/RSJ International Conference on Intelligent Robots and Systems}, 2011,
  pp. 1069--1074.

\bibitem{1389794}
S.~{Chernova} and M.~{Veloso}, ``An evolutionary approach to gait learning for
  four-legged robots,'' in \emph{2004 IEEE/RSJ International Conference on
  Intelligent Robots and Systems (IROS) (IEEE Cat. No.04CH37566)}, vol.~3,
  2004, pp. 2562--2567 vol.3.

\bibitem{Hengst2001OmnidirectionalLF}
B.~Hengst, D.~Ibbotson, S.~Pham, and C.~Sammut, ``Omnidirectional locomotion
  for quadruped robots,'' in \emph{RoboCup}, 2001.

\bibitem{Olave2003TheUR}
A.~Olave, D.~Wang, J.~Wong, T.~Tam, B.~Leung, M.~Kim, J.~Brooks, A.~Chang,
  N.~Huben, C.~Sammut, and B.~Hengst, ``The unsw robocup 2002 legged league
  team,'' \emph{RoboCup}, 01 2003.

\bibitem{article}
M.~S. Kim and W.~Uther, ``Automatic gait optimisation for quadruped robots,''
  in \emph{In Australasian Conference on Robotics and Automation}, 2003.

\bibitem{10.5555/1625275.1625428}
D.~Lizotte, T.~Wang, M.~Bowling, and D.~Schuurmans, ``Automatic gait
  optimization with gaussian process regression,'' in \emph{Proceedings of the
  20th International Joint Conference on Artifical Intelligence}, ser.
  IJCAI'07.\hskip 1em plus 0.5em minus 0.4em\relax San Francisco, CA, USA:
  Morgan Kaufmann Publishers Inc., 2007, p. 944–949.

\bibitem{6907117}
R.~{Calandra}, A.~{Seyfarth}, J.~{Peters}, and M.~P. {Deisenroth}, ``An
  experimental comparison of bayesian optimization for bipedal locomotion,'' in
  \emph{2014 IEEE International Conference on Robotics and Automation (ICRA)},
  2014, pp. 1951--1958.

\bibitem{Cully_2015}
\BIBentryALTinterwordspacing
A.~Cully, J.~Clune, D.~Tarapore, and J.-B. Mouret, ``Robots that can adapt like
  animals,'' \emph{Nature}, vol. 521, no. 7553, p. 503–507, May 2015.
  [Online]. Available: \url{http://dx.doi.org/10.1038/nature14422}
\BIBentrySTDinterwordspacing

\bibitem{10.3389/fnbot.2021.627157}
\BIBentryALTinterwordspacing
W.~Ouyang, H.~Chi, J.~Pang, W.~Liang, and Q.~Ren, ``Adaptive locomotion control
  of a hexapod robot via bio-inspired learning,'' \emph{Frontiers in
  Neurorobotics}, vol.~15, p.~1, 2021. [Online]. Available:
  \url{https://www.frontiersin.org/article/10.3389/fnbot.2021.627157}
\BIBentrySTDinterwordspacing

\bibitem{phdthesis}
M.~Shahriari, ``Design, implementation and control of a hexapod robot using
  reinforcement learning approach,'' Ph.D. dissertation, 07 2013.

\bibitem{lele2020learning}
A.~S. Lele, Y.~Fang, J.~Ting, and A.~Raychowdhury, ``Learning to walk: Spike
  based reinforcement learning for hexapod robot central pattern generation,''
  2020.

\bibitem{RAMEZANIDOORAKI2021103671}
\BIBentryALTinterwordspacing
A.~{Ramezani Dooraki} and D.-J. Lee, ``An innovative bio-inspired flight
  controller for quad-rotor drones: Quad-rotor drone learning to fly using
  reinforcement learning,'' \emph{Robotics and Autonomous Systems}, vol. 135,
  p. 103671, 2021. [Online]. Available:
  \url{https://www.sciencedirect.com/science/article/pii/S092188902030511X}
\BIBentrySTDinterwordspacing

\bibitem{koch2018reinforcement}
W.~Koch, R.~Mancuso, R.~West, and A.~Bestavros, ``Reinforcement learning for
  uav attitude control,'' 2018.

\bibitem{8453468}
M.~B. {Vankadari}, K.~{Das}, C.~{Shinde}, and S.~{Kumar}, ``A reinforcement
  learning approach for autonomous control and landing of a quadrotor,'' in
  \emph{2018 International Conference on Unmanned Aircraft Systems (ICUAS)},
  2018, pp. 676--683.

\bibitem{6990997}
Y.~{Vaghei}, A.~{Ghanbari}, and S.~M. R.~S. {Noorani}, ``Actor-critic neural
  network reinforcement learning for walking control of a 5-link bipedal
  robot,'' in \emph{2014 Second RSI/ISM International Conference on Robotics
  and Mechatronics (ICRoM)}, 2014, pp. 773--778.

\bibitem{8793627}
G.~A. {Castillo}, B.~{Weng}, A.~{Hereid}, Z.~{Wang}, and W.~{Zhang},
  ``Reinforcement learning meets hybrid zero dynamics: A case study for
  rabbit,'' in \emph{2019 International Conference on Robotics and Automation
  (ICRA)}, 2019, pp. 284--290.

\bibitem{castillo2019hybrid}
G.~A. Castillo, B.~Weng, W.~Zhang, and A.~Hereid, ``Hybrid zero dynamics
  inspired feedback control policy design for 3d bipedal locomotion using
  reinforcement learning,'' 2019.

\bibitem{rajeswaran2018learning}
A.~Rajeswaran, V.~Kumar, A.~Gupta, G.~Vezzani, J.~Schulman, E.~Todorov, and
  S.~Levine, ``Learning complex dexterous manipulation with deep reinforcement
  learning and demonstrations,'' 2018.

\bibitem{long2018optimally}
P.~Long, T.~Fan, X.~Liao, W.~Liu, H.~Zhang, and J.~Pan, ``Towards optimally
  decentralized multi-robot collision avoidance via deep reinforcement
  learning,'' 2018.

\bibitem{10.1145/3072959.3073602}
\BIBentryALTinterwordspacing
X.~B. Peng, G.~Berseth, K.~Yin, and M.~Van De~Panne, ``Deeploco: Dynamic
  locomotion skills using hierarchical deep reinforcement learning,'' \emph{ACM
  Trans. Graph.}, vol.~36, no.~4, Jul. 2017. [Online]. Available:
  \url{https://doi.org/10.1145/3072959.3073602}
\BIBentrySTDinterwordspacing

\bibitem{10.1007/s10514-018-9697-6}
\BIBentryALTinterwordspacing
P.~Kormushev, B.~Ugurlu, D.~G. Caldwell, and N.~G. Tsagarakis, ``Learning to
  exploit passive compliance for energy-efficient gait generation on a
  compliant humanoid,'' \emph{Auton. Robots}, vol.~43, no.~1, p. 79–95, Jan.
  2019. [Online]. Available: \url{https://doi.org/10.1007/s10514-018-9697-6}
\BIBentrySTDinterwordspacing

\bibitem{Peng_2018}
\BIBentryALTinterwordspacing
X.~B. Peng, P.~Abbeel, S.~Levine, and M.~van~de Panne, ``Deepmimic,'' \emph{ACM
  Transactions on Graphics}, vol.~37, no.~4, p. 1–14, Aug 2018. [Online].
  Available: \url{http://dx.doi.org/10.1145/3197517.3201311}
\BIBentrySTDinterwordspacing

\bibitem{wu2019investigation}
C.-A. Wu, ``Investigation of different observation and action spaces for
  reinforcement learning on reaching tasks,'' 2019.

\end{thebibliography}
\end{document}